\documentclass[10pt,twocolumn,letterpaper]{article}

\usepackage{cvpr}
\usepackage{times}
\usepackage{epsfig}
\usepackage{graphicx}
\usepackage{amsmath}
\usepackage{amssymb}
\usepackage{nicefrac}
\usepackage{mathtools}
\usepackage{adjustbox}

\usepackage{caption}
\usepackage{tikz}
\usepackage{pgfplots}
\usetikzlibrary{spy,calc}
\usepackage{array}
\newcolumntype{x}[1]{>{\centering\arraybackslash\hspace{0pt}}p{#1}}

\newif\ifblackandwhitecycle
\gdef\patternnumber{0} 

\pgfkeys{/tikz/.cd,
    zoombox paths/.style={
        draw=orange,
        very thick
    },
    black and white/.is choice,
    black and white/.default=static,
    black and white/static/.style={ 
        draw=white,   
        zoombox paths/.append style={
            draw=white,
            postaction={
                draw=black,
                loosely dashed
            }
        }
    },
    black and white/static/.code={
        \gdef\patternnumber{1}
    },
    black and white/cycle/.code={
        \blackandwhitecycletrue
        \gdef\patternnumber{1}
    },
    black and white pattern/.is choice,
    black and white pattern/0/.style={},
    black and white pattern/1/.style={    
            draw=white,
            postaction={
                draw=black,
                dash pattern=on 2pt off 2pt
            }
    },
    black and white pattern/2/.style={    
            draw=white,
            postaction={
                draw=black,
                dash pattern=on 4pt off 4pt
            }
    },
    black and white pattern/3/.style={    
            draw=white,
            postaction={
                draw=black,
                dash pattern=on 4pt off 4pt on 1pt off 4pt
            }
    },
    black and white pattern/4/.style={    
            draw=white,
            postaction={
                draw=black,
                dash pattern=on 4pt off 2pt on 2 pt off 2pt on 2 pt off 2pt
            }
    },
    zoomboxarray inner gap/.initial=5pt,
    zoomboxarray columns/.initial=2,
    zoomboxarray rows/.initial=2,
    subfigurename/.initial={},
    figurename/.initial={zoombox},
    zoomboxarray/.style={
        execute at begin picture={
            \begin{scope}[
                spy using outlines={%
                    zoombox paths,
                    width=\imagewidth / \pgfkeysvalueof{/tikz/zoomboxarray columns} - (\pgfkeysvalueof{/tikz/zoomboxarray columns} - 1) / \pgfkeysvalueof{/tikz/zoomboxarray columns} * \pgfkeysvalueof{/tikz/zoomboxarray inner gap} -\pgflinewidth,
                    height=\imageheight / \pgfkeysvalueof{/tikz/zoomboxarray rows} - (\pgfkeysvalueof{/tikz/zoomboxarray rows} - 1) / \pgfkeysvalueof{/tikz/zoomboxarray rows} * \pgfkeysvalueof{/tikz/zoomboxarray inner gap}-\pgflinewidth,
                    magnification=3,
                    every spy on node/.style={
                        zoombox paths
                    },
                    every spy in node/.style={
                        zoombox paths
                    }
                }
            ]
        },
        execute at end picture={
            \end{scope}
     \gdef\patternnumber{0}
        },
        spymargin/.initial=0.5em,
        zoomboxes xshift/.initial=1,
        zoomboxes right/.code=\pgfkeys{/tikz/zoomboxes xshift=1},
        zoomboxes left/.code=\pgfkeys{/tikz/zoomboxes xshift=-1},
        zoomboxes yshift/.initial=0,
        zoomboxes above/.code={
            \pgfkeys{/tikz/zoomboxes yshift=1},
            \pgfkeys{/tikz/zoomboxes xshift=0}
        },
        zoomboxes below/.code={
            \pgfkeys{/tikz/zoomboxes yshift=-1},
            \pgfkeys{/tikz/zoomboxes xshift=0}
        },
        caption margin/.initial=0.4ex,
    },
    adjust caption spacing/.code={},
    image container/.style={
        inner sep=0pt,
        at=(image.north),
        anchor=north,
        adjust caption spacing
    },
    zoomboxes container/.style={
        inner sep=0pt,
        at=(image.north),
        anchor=north,
        name=zoomboxes container,
        xshift=\pgfkeysvalueof{/tikz/zoomboxes xshift}*(\imagewidth+\pgfkeysvalueof{/tikz/spymargin}),
        yshift=\pgfkeysvalueof{/tikz/zoomboxes yshift}*(\imageheight+\pgfkeysvalueof{/tikz/spymargin}+\pgfkeysvalueof{/tikz/caption margin}),
        adjust caption spacing
    },
    calculate dimensions/.code={
        \pgfpointdiff{\pgfpointanchor{image}{south west} }{\pgfpointanchor{image}{north east} }
        \pgfgetlastxy{\imagewidth}{\imageheight}
        \global\let\imagewidth=\imagewidth
        \global\let\imageheight=\imageheight
        \gdef\columncount{1}
        \gdef\rowcount{1}
        
    },
    image node/.style={
        inner sep=0pt,
        name=image,
        anchor=south west,
        append after command={
            [calculate dimensions]
            node [image container,subfigurename=\pgfkeysvalueof{/tikz/figurename}-image] {\phantomimage}
            node [zoomboxes container,subfigurename=\pgfkeysvalueof{/tikz/figurename}-zoom] {\phantomimage}
        }
    },
    color code/.style={
        zoombox paths/.append style={draw=#1}
    },
    connect zoomboxes/.style={
    spy connection path={\draw[draw=none,zoombox paths] (tikzspyonnode) -- (tikzspyinnode);}
    },
    help grid code/.code={
        \begin{scope}[
                x={(image.south east)},
                y={(image.north west)},
                font=\footnotesize,
                help lines,
                overlay
            ]
            \foreach \x in {0,1,...,9} { 
                \draw(\x/10,0) -- (\x/10,1);
                \node [anchor=north] at (\x/10,0) {0.\x};
            }
            \foreach \y in {0,1,...,9} {
                \draw(0,\y/10) -- (1,\y/10);                        \node [anchor=east] at (0,\y/10) {0.\y};
            }
        \end{scope}    
    },
    help grid/.style={
        append after command={
            [help grid code]
        }
    },
}

\newcommand\phantomimage{%
    \phantom{%
        \rule{\imagewidth}{\imageheight}%
    }%
}
\newcommand\zoombox[2][]{
    \begin{scope}[zoombox paths]
        \pgfmathsetmacro\xpos{
            (\columncount-1)*(\imagewidth / \pgfkeysvalueof{/tikz/zoomboxarray columns} + \pgfkeysvalueof{/tikz/zoomboxarray inner gap} / \pgfkeysvalueof{/tikz/zoomboxarray columns} ) + \pgflinewidth
        }
        \pgfmathsetmacro\ypos{
            (\rowcount-1)*( \imageheight / \pgfkeysvalueof{/tikz/zoomboxarray rows} + \pgfkeysvalueof{/tikz/zoomboxarray inner gap} / \pgfkeysvalueof{/tikz/zoomboxarray rows} ) + 0.5*\pgflinewidth
        }
        \edef\dospy{\noexpand\spy [
            #1,
            zoombox paths/.append style={
                black and white pattern=\patternnumber
            },
            every spy on node/.append style={#1},
            x=\imagewidth,
            y=\imageheight
        ] on (#2) in node [anchor=north west] at ($(zoomboxes container.north west)+(\xpos pt,-\ypos pt)$);}
        \dospy
        \pgfmathtruncatemacro\pgfmathresult{ifthenelse(\columncount==\pgfkeysvalueof{/tikz/zoomboxarray columns},\rowcount+1,\rowcount)}
        \global\let\rowcount=\pgfmathresult
        \pgfmathtruncatemacro\pgfmathresult{ifthenelse(\columncount==\pgfkeysvalueof{/tikz/zoomboxarray columns},1,\columncount+1)}
        \global\let\columncount=\pgfmathresult
        \ifblackandwhitecycle
            \pgfmathtruncatemacro{\newpatternnumber}{\patternnumber+1}
            \global\edef\patternnumber{\newpatternnumber}
        \fi
    \end{scope}
}

\usepackage[acronym,shortcuts]{glossaries}

\usepackage{setspace}

\usepackage{booktabs}

\cvprfinalcopy 

\def\cvprPaperID{1948} 

\ifcvprfinal\fi

\usepackage{subcaption}

\begin{document}

\title{Convolutional Neural Networks Analyzed via Inverse Problem Theory and Sparse Representations}

\onehalfspacing
\author{Cem Tarhan, Gozde Bozdagi Akar, \\
        Middle East Technical University\\
        {\tt\footnotesize \{cem.tarhan,bozdagi\}@metu.edu.tr}
        }

\maketitle

\singlespacing
\begin{abstract}
    Inverse problems in imaging such as denoising, deblurring, superresolution (SR) have been addressed for many decades.  In recent years, convolutional neural networks (CNNs) have been widely used for many inverse problem areas. Although their indisputable success, CNNs are not mathematically validated as to how and what they learn. In this paper, we prove that during training, CNN elements solve for inverse problems which are optimum solutions stored as CNN neuron filters. We discuss the necessity of mutual coherence between CNN layer elements in order for a network to converge to the optimum solution. We prove that required mutual coherence can be provided by the usage of residual learning and skip connections. We have set rules over training sets and depth of networks for better convergence, i.e. performance.\footnote{This paper is a postprint of a paper submitted to and accepted for publication in IET Signal Processing Journal and is subject to Institution of Engineering and Technology Copyright. The copy of record is available at the IET Digital Library}
\end{abstract}

\section{Introduction}
\label{S:1}

In many image processing applications, an observational system model is the initial step to a solution. An observation g can be modeled as an output of a system K(.) when stimulated by an input t. Finding the variable t for an observation g and system K is named as an inverse problem. Due to the nature of observation modelling in imaging applications, the inversion of system K is ill-conditioned. Small variations in the observed value causes huge swings in estimated input value. This perturbation is caused by the noise in the system which is denoted by n.

\begin{equation}\label{eq_1} g = Kt + n \end{equation}

Many areas of research deal with inverse problems under the name of denoising, deblurring, deconvolution, enhancement, restoration, single image superresolution, MRI and CT reconstruction from projections etc. These problems can be classified as inverse problems that are defined with convolution operations. For decades, analytical methods have been utilized for solving these problems. A deterministic approach is to minimize a data fidelity term 

\begin{equation}\label{eq_a}\ \hat{t} = \underset{t}{argmin}\{\Delta (t,K,g)\} \end{equation}

The $\Delta(.)$ can be $||g-Kt||^2$ to obtain a Least Squares (LS) solution . Other methods include $L_P$ norm solution, Kullback Liebler (KL) distance solution. Since this problem is ill-conditioned a regularization term can be added to the data fidelity term. Regularization, r(t), adds a mismatched function to the cost function that will balance the solution. 

\begin{equation}\label{eq_3} \underset{t}{argmin} \{\Delta (t,K,g) + r(t) \} \end{equation}

Statistical inversion methods are another approach for an inverse problem. The observation vector \textbf{g} is assumed to be drawn from a probabilistic distribution p(\textbf{g}). For this approach, the cost function will become $\Delta (t,g) = p(g|t)$ for a maximum likelihood approach. Stochastic approaches are out of scope for this work, therefore related literature is not included.

Analytical methods seek to successfully formulate the system model and find the optimum solution strategy with proper regularizations. These methods require careful mathematical analysis and costly solutions. 

Another method that has been successfully used in inverse problem solutions is neural network approach, specifically convolutional neural networks (CNNs) for image processing applications. CNNs are trained by a set of images to learn a mapping from g to t. The structure of the network; number of layers and elements in each layer, interconnections between layers, special layers such as pooling or normalization changes depending on the specific application. However the learning process is generally based on stochastic gradient descent type coefficient updates. In most cases the cost function is chosen as mean square error (MSE). In some cases, such as generative adversarial networks and variational autoencoders, instead of end to end mapping from g to t, some feature discrimination is applied before calculating an MSE value together with a cross entropy loss function.

Recent advances on technology enabled training of bigger networks therefore providing better results \cite{LECUN}\cite{UNSER}\cite{KATSAGGELOS}. A sample of methods available in literature, which we have included in the Background part, show that CNN performance surpasses that of analytical methods in all aspects. Then one could ask "what is ahead of us?" The answer would be understanding how a network learns and how it uses the learned parameters. The former question is about the training phase and the latter is about the testing phase. The testing phase, the forward pass, can be shown to be satisfying a set of equations for application specific cases and setups. Unfolded iterative shrinkage thresholding (IST) iterations \cite{GREGOR}, alternating direction method of multipliers (ADMM) iterations are calculated by network layers \cite{ADMMNET} and layer outputs are visualized as sparse representation vectors \cite{SRCNN} or higher dimension manifold projections \cite{MANIFOLD}. The training phase is not addressed as often in the literature to answer what a CNN actually solves during backpropagation. 

In order to be able to analyze a complex structure such as CNN with mathematically tractable methods, we apply some simplifications and then generalize the results to fully operational network.

To this end we can summarize the contributions of this work as:
\begin{itemize}
    \item We show that CNN elements, i.e. neuron filters, solve for an inverse problem during training. The operator of the inverse problem is taken as a dictionary matrix constructed from input training data. The solution of the problem are the neuron filter coefficients. These coefficients are representation vectors of the target, which act similarly as described in manifold theory \cite{MANIFOLD} where each neuron filter acts as a point in high dimensional space to form a manifold for a solution. Differently from the argument in the literature \cite{LECUN} \cite{UNSER} \cite{KATSAGGELOS} that trained networks resemble or become inverse problem solvers, we claim that the training process is carried out in this fashion.
    \item By using \textit{Representation Dictionary Duality} (RDD) we relate sparse representation and convolutional sparse coding \cite{PAPYAN16} to the evolution of CNNs and provide a mathematical basis for skip connections \cite{MAO} and residual learning \cite{VDSR}.
    \item We propose a method to separate the training set for maximizing the efficiency of training and number of features (filters) which is consistent with mathematical foundations that we set. Also we suggest a method to decide on the depth and structure of the CNN.
\end{itemize}

Rest of the paper is organized as follows: in section 2 we refer to a vast literature about CNNs used for various inverse problems and analytical approaches for inverse problems. In section 3 we provide mathematical proofs for CNN training as inverse problem solutions and necessity of skip connections. In section 4 we have demonstrated experimental validations for key theoretical points. In section 5 we summarize our contributions and point out to further research areas.

\section{Background}
\label{S:2}

As stated before, there are different approaches to find the solution to the inverse problem given in equation \ref{eq_a}. In the rest of this section we will describe in detail the methods that address inverse problems.








\subsection{Data Driven Approaches and CNNs}

Before discussing the analytical methods that approach inverse problems with aid of mathematical modelling and iterative solutions we discuss the literature on Data Driven Approaches (DDA). Instead of approaching the inverse problem to directly invert an observation model, DDA learn mappings from input and target training images. Methods either learn a compact dictionary \cite{YANG08}\cite{YANG10}\cite{ANR}\cite{A+} or train a model that fits the problem and learn parameters for the model \cite{RAISR}\cite{TNRD}.

Dictionary based DDA jointly solve for a compact dictionary and a representation vector. Sparse representation has been applied to the dictionary learning based SR problem where an input image is sparsely represented by a dictionary. The representation vector is applied to another library for reconstruction of output image. These algorithms both solve for creating dictionary and solve for a representation vector for any input. 

The K-SVD algorithm \cite{KSVD} is one of the keystones of dictionary learning for the purpose of sparse representation. Aharon et. al. have proposed the usage of a compact dictionary D, from which a set of atoms (columns or dictionary elements) are to be selected via a vector f and the combination of these atoms is constrained to be similar to an observation patch (or image) g via $||g-Df||_p\le\varepsilon$. If the dimension of g is less than that of columns of matrix D and if D is full-rank matrix then there are infinitely many solutions to the problem therefore a sparsity constraint is introduced.

\begin{equation}
\label{eq_12} \underset{f}{\min}  \ || f ||_0 \ s.t. \  || g- Df ||_2  \le \varepsilon 
\end{equation}
The L$_0$ norm gives the number of entries in f that are non-zero. 

The usage of compact dictionaries for SR problem is introduced in \cite{YANG08}. The authors have used the approach of K-SVD to learn representation vectors and dictionaries. Instead of using an $L_0$ normed regularization and $L_1$ normed regularization is used which still guaranties sparsity for the formulated problem. Such a problem is formulated using Lagrange multipliers.
\begin{equation}
\label{eq_13} 
\underset{f}{\min} \ \lambda||f||_1+\frac{1}{2}||g- Df||_2^2
\end{equation}

During learning phase the library D is initialized by random gaussian noise and an iterative algorithm between a batch representation matrix Z and dictionary D refines the dictionary while maintaining sparsity for representation vectors of training set. The estimation of sparse representation vectors are done by using basis pursuit methods. \cite{YANG08} uses two dictionaries, one for low resolution (LR) representation, one for high resolution (HR) reconstruction.

Timofte et. al. \cite{ANR} have proposed the usage of L$_2$ norm instead of L$_1$ norm for even faster computations. Although usage of L$_2$ norm eliminated the sparsity constraint from the equation, it will play a role in understanding how CNNs work in later sections.

Trainable models are also used for inverse problems. A method, trainable nonlinear reaction diffusion (TNRD) \cite{TNRD}, uses a training set to learn a set of filters and non-linearities that are applied in a cascaded fashion to the input image. The resulting algorithm uses stochastic gradient descent type of learning. The minimized objective function uses a suitable proximal mapping function among previously inspected functions \cite{COMBETTES}. In \cite{TNRD}, input image is filtered by a number of trained filters. Filtered images are then applied to a non-linear proximal function and then the output of non-linearities are further filtered by the same number of filters. The result is obtained by summing all the result. The approach of TNRD is quite similar to how a CNN operates although the model is not weaved in a network fashion since it lacks the hidden layers which give CNNs its power to learn highly nonlinear surfaces (manifolds). 

Another trainable model type method is proposed by Romano et. al. rapid and accurate image super resolution (RAISR) \cite{RAISR}. The algorithm learns a set of filters for training patches that are composed of differently oriented edges with various strength and coherencies. Separation of different content patches is carried out according to gradient information of input images. Even though RAISR uses a set of filters in a convolutional structure, the algorithm does not constitute a network structure.


In the last decade the usage of CNNs for inverse problems have dramatically increased, mainly due to advancing technology.

One of the first applications of inverse problems on CNNs was proposed by Gregor \& LeCun \cite{GREGOR} as learned iterative shrinkage thresholding algorithm (LISTA). The method deals with sparse coding problem that is formulated by \ref{eq_13}. Each layer of neural network type model of LISTA represents an unfolded iteration of IST type of solver that was inspected by Combettes et. al. as iterative soft-thresholding \cite{COMBETTES}. The forward pass of the algorithm approximates iterations of an IST algorithm. Training of LISTA is carried out in gradient descent type learning. Later Bronstein et. al. \cite{BRONSTEIN} have addressed the same algorithm and showed that the resulting layers of LISTA are not mere approximations to an IST algorithm but fully functional sparse encoders in their own right.

In \cite{MANIFOLD} authors have described representation learning as a manifold learning for which a higher dimensional data is represented compactly in a lower dimensional manifold. They have discussed that the variations in the input space is captured by the representations and each element of a network represents a higher dimensional coordinate point of the manifold. In the same paper authors have discussed the challenge of training deep networks; learning dynamics, convergence to good minimas of the cost function. The training of neural networks is not mathematically understood in current literature \cite{MANIFOLD}.


Schuler et. al. \cite{SCHULER} have proposed to unroll the recovery steps required for a deblurring operations with a neural network.




Mao et. al. \cite{MAO} have proposed the usage of symmetric convolutional-deconvolutional layers, hoping that the convolutional layers will encode the important features of the image while rejecting defects and deconvolutional layers will reconstruct the image without the defects. Their experiments have shown that proposed architecture yields better results compared to sole convolutional architectures. The authors have also used skip connections between convolutional and deconvolutional layers, expecting that it will cope with gradient vanishing problem for deeper networks. 

Zhang et. al. \cite{ZHANG} have proposed a denoising network by exploiting the similarities of execution of denoising CNNs to TNRD equations. 


Yang et. al. \cite{ADMMNET} have proposed a network called ADMM-Net for optimizing a compressive sensing reconstruction problem, taking on from the iterative steps of ADMM algorithm \cite{ADMM}. Each step of the network is weaved accoring to split augmented lagrangian solver approach. This approach separates the method from many methods in literature where a CNN structure is used for various inverse problem solutions without mathematical reasoning. 


The mapping between the high resolution (HR) and low resoultion (LR) images can also be found by convolutional networks for SR problem (\cite{SRCNN}, \cite{VDSR}).

The activation function plays an important role in neural network training. In many state-of-the-art algorithms major functions such as tanh and softmax have been replaced by rectified linear units \cite{RELU} that are linear approximations of mathematically complex and computationally heavy functions. Glorot et. al. \cite{GLOROT} have empirically shown that by using rectified activations, the network can learn sparse representations easier. For a given input, only a subset of hidden neurons are activated, leading to better gradient backpropagation for learning and better representations during forward pass. Especially sparse representation has been shown to be useful \cite{GLOROT}. Sparsity constraint provides information disentagling which allows the representation vectors to be robust against small changes in input data.

Dong et. al. \cite{SRCNN} have provided the earliest relation of CNNs to Sparse Representation. In their view outputs of the first layer of neurons constitute a representation vector for a patch around each pixel in LR image, second layer maps LR representations to HR representation vectors and the last layer reconstructs HR image using 5x5 sized filters (or atoms if we have used the jargon of sparse representations). Although this idea qualitatively maps CNNs as a solution method for sparse representation problem, in section \ref{S:3} we will show a more complete understanding with mathematical background. 

Bruna et. al. \cite{BRUNA} have used CNN for extraction of LR representation which would be used to reconstruct a proper HR image that is picked according to a gibbs distribution. Kim et. al. \cite{VDSR} have proposed usage of deep networks for SR problem and their unique approach of using residual learning helped the large network to converge in a reasonable time. 

In many SR applications LR image that is fed into the network is bicubically upsampled which reduces the SR problem into deconvolution (deblurring) problem. Shi et. al. \cite{ESPCN} have proposed the usage of CNN itself to learn upsample filters together with enhancement filters, therefore it was the first time an SR problem is completely addressed by CNN based algorithm.

In a recent paper, Papyan et. al. \cite{PAPYAN16} have discussed the connection of convolutional sparse coding and CNNs. By inspecting the activation of each layer of a trained network, it is proven that CNN layers become sparse representation dictionaries. In section \ref{S:3} we will use core ideas from \cite{PAPYAN16} to prove necessity of skip connections.

The literature on inverse problem related CNNs is vast. But many methods lack the proper mathematical explanation for their success. Before moving onto mathematical analysis of CNNs we will look into the analytical approaches for inverse problems in literature.

\subsection{Analytical Approach to Solution}
\label{AA}
The analytical approach for solving the inverse problem involves minimizing a non-smooth convex cost function with additional regularizations that are impinged on the optimization to statistically tie the solution to the observations. Mathematically, most imaging problems such as debluring, denoising, SR are most appropriately formulated with variational equations. The objective of these equations is to incorporate a priori information and enforces coherency with the observations. Equation \ref{eq_3} is an example of a variational equation where r(t) term is continuous and differentiable function.

The choice of r(t), the regularizer, depends on multiple parameters, such as the objective of the problem, the probability distribution of noise etc. For the case when the noise probability distribution is Gaussian $L_2$ norm is used as r(t) = $||t||^2$. The minimizer for equation \ref{eq_3} is given by the so called Landweber equation

\begin{equation}\label{eq_5} \hat{t}=(K^TK+\lambda I)^{-1}K^Tg \end{equation}

where $I$ is identity matrix and $\lambda$ is the regularization parameter that is chosen according to the application. Although derivation of quadratic functions is easier especially for gradient descent type solvers, they are known to stuck to local minima occasionally which is a major problem for imaging applications where the cost function is almost always multimodal. 

One of well known methods that address the solution of variational equations is alternating direction method of multipliers (ADMM) \cite{ADMM} . The method uses augmented lagrangian to split the optimization problem into two parts that are optimized in alternating steps. Later Combettes \& Wajs \cite{COMBETTES} have analyzed the convex optimization problems and their solutions by proximal forward-backward splitting. They have inspected various proximity mapping functions for solutions of inverse problems with projection onto convex sets. Linear inverse problem solution uses iterations which result in so called Iterative Shrinkage/Thresholding (IST) \cite{COMBETTES}. For linear inverse problems the solution to the inversion of equation \ref{eq_1} can be obtained by the help of Moreau proximity operator \cite{COMBETTES} as

\begin{equation}
\label{eq_6} 
t^n=prox_{b||.||}(t^{n-1}+b.K^T(g-Kt^{n-1}))
\end{equation}

where a class of proximity operators are defined, and n stands for iteration count. The special function for the case of $L_1$ regularization is soft thresholding function also known as shrinkage operator. Soft thresholding function for scalar input is defined in equation \ref{eq_7} where $b$ is the bias value. For vector inputs the same operation is applied to each element separately. 

\begin{equation}
\label{eq_7} 
prox_{b||.||}t=sign(t).\max(|t|-b,0)\doteq S_{b}(t)
\end{equation}

The funcion S$_b$ is symmetric soft thresholding, as we shall see later; and it has an inherent connection to non-negative soft thresholding used in neural networks. Notice that $K^T(g-Kt^{n-1})$ is the negative gradient of data fidelity term in the original formulation. Using IST iterations, the solution for the inverse problem is obtained iteratively. Each iteration uses an update term that uses gradient values thresholded by Moreau proximity mapping. This method is also named as Proximal Landweber Iterations \cite{DAUBECHIES}. 

Daubechies et. al.  \cite{DAUBECHIES}  have proposed the usage of non-quadratic regularization constraints that promote sparsity by the help of an orthonormal (or overcomplete) basis {$\varphi_l$} of a Hilbert space. For the problem defined in equation \ref{eq_3} it is proposed to use regularization function r(t) as

\begin{equation}
\label{eq_8} 
r(t)=\sum_{\forall l}b_l|\langle t,\varphi_l\rangle|^p 
\end{equation}

where $b_l$ are coefficients. Usage of this regularization instead of $L_1$ norm provides a better representation for information that has space-varying smoothness property, such as real life images. For the case when p = 1, a straightforward variational equation can be obtained in an iterative way for equation \ref{eq_3}. 
\begin{equation}
\label{eq_9} 
\langle t^n,\varphi_l\rangle=S_b(\langle t^{n-1},\varphi_l\rangle+\langle K^T(g-Kt^{n-1}),\varphi_l\rangle)
\end{equation}

Iterations over the set of basis functions can be carried out in one formula
\begin{equation}
\label{eq_10} 
t^n=Z_b(t^{n-1}+K^T(g-Kt^{n-1}))
\end{equation}
where
\begin{equation}
\label{eq_11} 
Z_b(x)\doteq\sum_{l\in\Gamma}S_b(\langle x,\varphi_l\rangle)\varphi_l
\end{equation}
which can be seen as a method to file the elements of x in the direction of $\varphi_l$. Daubechies et. al. have proven that the solution obtained by iterating t reaches the global minimum of the solution space and solution method is stable. The solution will reach to an optimum point if K is a bounded operator satisfying $||Kt|| \le C||t||$ for any vector t and some constant C.

Although more research has been put into regularization methods later starting with Combettes \& Wajs \cite{COMBETTES}, the method of Daubechies et. al. is suitable for us to explain training of CNNs later.

\section{Mathematical Analysis of CNNs}
\label{S:3}
It is as important to know how a CNN learns as much as what a trained CNN represents or resembles. The latter is discussed in literature for many different areas of application, the former is lacking proper explanation and mathematical proving. The main reason is that a mathematical model for a CNN cannot be obtained explicitly, the fastest method to analyze how a CNN learns is to let it learn. For this reason we start from the very beginning and formulate CNN operations for simplified cases. We show the optimality of CNN elements and conditions for optimality.
\subsection{Training (Learning) Phase}

For the training phase of CNNs, input images are fed into the network for forward pass. The resulting image from the network is compared against a ground truth (GT) image and the error is backpropagated. Since the input image is convolved by the neuron filter, its size should be larger than the size of the output to prevent boundary conditions. It is reported that boundary conditions do not cause a trouble in a residual learning environment \cite{VDSR}.

\begin{figure}
	\centering\includegraphics[width=1\linewidth]{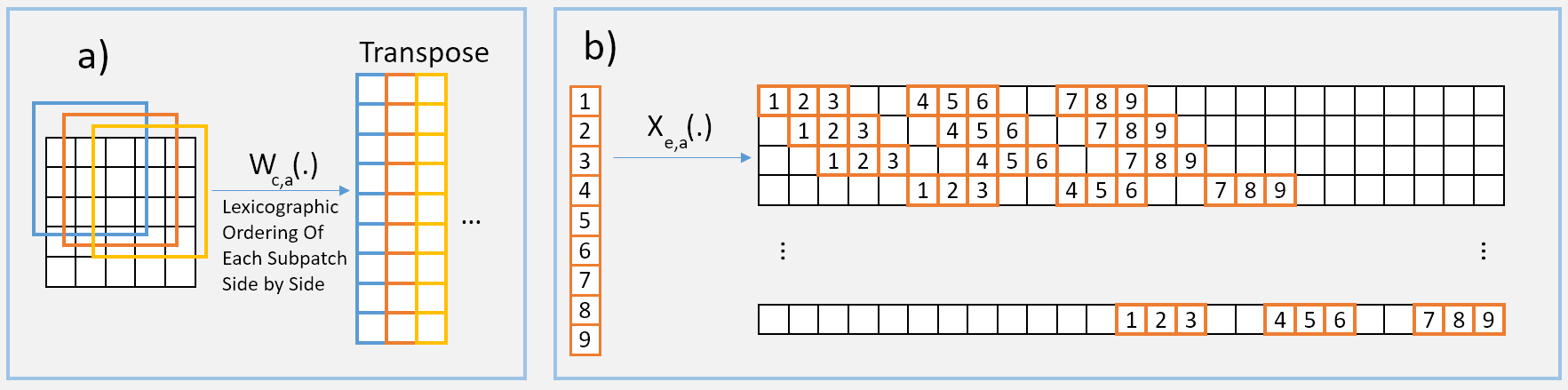}
	\caption{Operators: a) W(.) operator for lexicographical dictionary creation from a patch, b) X(.) operator for equivalent operation for an order swapping of cascaded convolutions}
	\label{XOPERATOR}
\end{figure}

Before moving on to explaining CNN operations we have to define a few operators. These operators will be useful in describing convolutional operations with algebraic equations. Take a filter, \textit{f}, of size $a\text{x}a$; an input patch, $x_{k-1}$, of size $c\text{x}c$. The valid part of the output, $x_{k}$, of convolution of \textit{f} and $x_{k-1}$ will have a size of $(c-a+1)\text{x}(c-a+1)$.

\begin{equation*}
x_{k} =  x_{k-1}*f 
\end{equation*}

where $*$ is convolution operation. To show these operations algebraically we define a new operator W$_{c,a}$(.) which takes smaller patches (subpatches) from a larger patch (superpatch), orders them in lexicographical order then concatenates all vectors into a matrix. This operator is depicted in Figure \ref{XOPERATOR} a. Subscripts indicate the sizes of $x_{k-1}$ and \textit{f} respectively as $c$x$c$ and $a$x$a$. The resulting matrix from W$_{c,a}$(.) operator will have a size of $(c-a+1)^2 \  \text{x}\  a^2$.


\begin{equation}
\boldsymbol{x_{k-1}*f}=W_{c,a}(x_{k-1}).\boldsymbol{f}
\end{equation}

\begin{figure}
	\centering\includegraphics[width=1\linewidth]{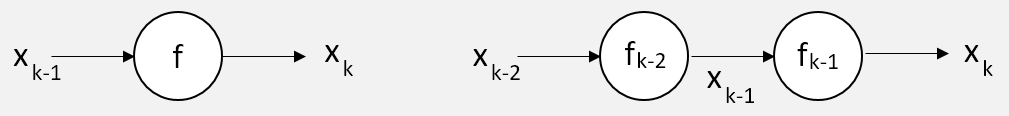}
	\caption{a) One neuron filter b) Two cascaded neuron filters}
	\label{ONE_TWO}
\end{figure}

We denote lexicographically vectorized 2-D patches with bold characters. Notice in this case result of convolution $x_{k-1}*f$ is also vectorized. We also define an X$_{e,a}$(.) operator to denote an order swapping between two cascaded convolutions as shown in Figure \ref{XOPERATOR} b. The index a indicates the size of the filter, $a\text{x}a$, that is given as input to the operator. The index e indicates the size of the image patch, $e\text{x}e$, with which the input filter is going to be convolved. Take an input $x_{k-2}$ as in Figure \ref{ONE_TWO} b, with size $e\text{x}e$, and two filters \textit{f}$_{k-2}$ and \textit{f}$_{k-1}$ with sizes $a\text{x}a$ the resulting operations will be as follows

\begin{align}
x_{k-2}*f_{k-2}*f_{k-1} &= x_{k-2}*f_{k-1}*f_{k-2}
\\ &=X_{e,a}(f_{k-1}).W_{e,a}(x_{k-2}).\boldsymbol{f_{k-2}}\notag
\end{align}

To start analyzing a CNN we are going to take one CNN element, a neuron filter into consideration depicted in Figure \ref{ONE_TWO} a. 

\begin{equation}
x_k =  soft_b(x_{k-1}*f)  
\end{equation}

where $soft_b$, nonnegative soft threshdoling, is defined for each scalar input as in equation \ref{eqsoft}. For vector inputs same operations are applied to each element in the vector. The output of a neuron filter is outcome of nonnegative soft thresholding applied to the convolution result.
\begin{equation}
\label{eqsoft}
soft_b(x) \doteq max(x-b,0)
\end{equation}

During training a target image, $t$, is used to calculate an error. The mean square error is used for gradient calculation and parameter updates. We define a dictionary for learning (training) phase as
\begin{align}
\label{oops}
D_{L,k-1} &\doteq W_{c,a}(x_{k-1}) 
\\\boldsymbol{x_{k}} &= soft_b(D_{L,k-1}\boldsymbol{f}) 
\\error &= \boldsymbol{t} - soft_b(D_{L,k-1}\boldsymbol{f})\label{oops2}
\\mse &= \frac{1}{2}||\boldsymbol{t}-soft_b(D_{L,k-1}\boldsymbol{f})||^2
\\\frac{\partial mse}{\partial \boldsymbol{f}} &=-(D_{L,k-1}^T (\boldsymbol{t} - soft_{\boldsymbol{b}}(D_{L,k-1} \boldsymbol{f})))
\end{align}

Moving forward with CNN learning operations, parameter updates are carried out by adding negative gradient on top of old value. 
\begin{align}
\boldsymbol{f^{n}}&=\boldsymbol{f^{n-1}}-\frac{\partial mse}{\partial \boldsymbol{f^{n-1}}} 
\\&= \boldsymbol{f^{n-1}} + D_{L,k-1}^T (\boldsymbol{t} - soft_{\boldsymbol{b}}(D_{L,k-1} \boldsymbol{f^{n-1}})) \label{lricin}
\end{align}
where n is the iteration number. In Daubechies et. al. \cite{DAUBECHIES} (Remark 2.4) the learning rate was introduced with a matrix, G, that is diagonal in $\varphi_l$ basis as $G \varphi_l = \eta_l \varphi_l$. The term $\eta_l$ modifies equation \ref{lricin} as 
\begin{equation}
\boldsymbol{f^{n}} = \boldsymbol{f^{n-1}} + \frac{1}{\eta_l} D_{L,k-1}^T (\boldsymbol{t} - soft_{\boldsymbol{b}}(D_{L,k-1} \boldsymbol{f^{n-1}}))
\end{equation}

Daubechies et. al. have commented that the solution can be followed by this equation but they have omitted it for simplicity of equations. It will affect the convergence speed in general but it was proven that the solution reaches to optimum point either way. Since we show that CNN gradient descent based learning yields the same solution as Daubechies' solution we also omit this term and use equation \ref{lricin}.

The $f^n$ can be defined piece-wise as
\begin{equation}
f^n = \left\{
        \begin{array}{ll}
            {f^{n-1}} + D_{L,k-1}^T (t -  D_{L,k-1} f^{n-1} + b) & \\ \quad \quad  \quad  \quad \quad \quad \quad  \quad  \quad \quad \quad \quad \quad  \quad   if \ D_{L,k-1} {f^{n-1}} > b \\
            {f^{n-1}} + D_{L,k-1}^T {t} & \\ \quad \quad  \quad  \quad \quad \quad \quad  \quad  \quad \quad \quad \quad \quad  \quad  if \ D_{L,k-1} {f^{n-1}} \leq b
        \end{array}
    \right.
\end{equation}
We want to collect the iterations inside the soft thresholding to modify equations towards Daubechies' solution.
\begin{equation}
\label{gradient}
f^n = soft_{{b'}}({f^{n-1}} + D_{L,k-1}^T({t}-D_{L,k-1} {f^{n-1}}))
\end{equation}

To have to $f^n$ definitions to be equal then we would have to describe b' as
\begin{equation}
b' = \left\{
        \begin{array}{ll}
            -D_{L,k-1}^T {b} & \quad D_{L,k-1} {f^{n-1}} > b \\
            -D_{L,k-1}^T D_{L,k-1} {f^{n-1}} & \quad D_{L,k-1} {f^{n-1}} \leq b
        \end{array}
    \right.
\end{equation}

Although this split seems to describe four zoned thresholding function, this is not the case. Since we are dealing with natural images $D_{L,k-1}$ and $t$ have positive values. This reduces the thresolding function to two zones again. Another remark is that we did not use bold letters for f,t and b, which indicates that we assumed scalar f,t and b. This is a simplification to avoid using indexes in an already complicated formula. Also it is important to emphasize that we are not proposing new functions for the training of a neuron. We are changing the equations to make them tractable.

For simplicity define
\begin{equation}
\boldsymbol{e^n} \doteq \boldsymbol{f^{n-1}} + D_{L,k-1}^T(\boldsymbol{t}-D_{L,k-1} \boldsymbol{f^{n-1}})
\end{equation}

Introduce $\{\varphi_l\}$ a CON basis vector set of Hilbert space.
\begin{equation}
\langle \boldsymbol{f^{n}},\varphi_l \rangle = \langle soft_{\boldsymbol{b}'}(\boldsymbol{e^n}),\varphi_l\rangle
\end{equation}

Now define
\begin{equation}
h^n \doteq soft_{b_l}(\langle \boldsymbol{e^n},\varphi_l \rangle)
\end{equation}

Explicitly
\begin{equation}
\langle \boldsymbol{f^{n}},\varphi_l \rangle = f^n[1]\varphi_l [1] +...+f^n [M] \varphi_l [M]
\end{equation}
\begin{equation}
h^n = e^n[1]\varphi_l [1] +...+e^n [M] \varphi_l [M] - b_l
\end{equation}

where M is the length of vector f. Since we actively control $b'$ values for each computation, we know that $soft_{\boldsymbol{b'}}$ is active all times, only ${b'}$ will change. Thus $f^n [i] = e^n [i] -  b'[i]$ at all times. Therefore to have $h^n$ equal to $\langle \boldsymbol{f^{n}},\varphi_l \rangle$ we should have  $b_l = \langle \boldsymbol{b'} , \varphi_l \rangle $.

\begin{align}
\label{eq_14}
\boldsymbol{f^{n}}&=\sum_{\forall l}\langle \boldsymbol{f^{n}},\varphi_l \rangle  \varphi_l =\sum_{\forall l}soft_{b_l}(\langle \boldsymbol{e^n},\varphi_l \rangle) \varphi_l \notag
\\ &=\sum_{\forall l}S_{b_l}(\langle \boldsymbol{f^{n-1}} + D_{L,k-1}^T( \boldsymbol{t}-D_{L,k-1} \boldsymbol{f^{n-1}}),\varphi_l \rangle)\varphi_l 
\end{align}

where $S_{b_l}$ is defined in equation \ref{eq_7}. Since the basis vectors can be chosen with two different directions without loss of generality we can choose $\varphi_l$ such that the innerproduct (or projection) always yields a nonnegative result. By doing so we can use symmetric soft thresholding ($S_b$) instead of nonnegative soft thresholding ($soft_b$). This enables CNN equations to become exact inverse problem solutions. We will replace operator K in equation \ref{eq_10} with  $D_{L,k-1}$ in equation \ref{eq_14}. From previous discussion we know that the operator needs to be bounded for the solution to exist for equation \ref{eq_10}. In this case $D_{L,k-1}$ matrix which is composed of image patches is bounded. \textbf{Therefore equation \ref{eq_14} shows that a neuron filter solves an inverse problem during training and that it is optimal and stable solution.} 

A neuron filter solves a system as in equation \ref{trainprob}

\begin{equation}
\label{trainprob} 
t = D_L f + n'
\end{equation}

where t is the target image as before, $D_L$ is the dictionary matrix constructed from input data and f is the neuron filter to be learned, n' is noise. The solution of such a system is given by equation \ref{eq_14}. Notice that the training step is an intermediary step of using a CNN for the actual inverse problem of equation \ref{eq_1}. As we will discuss in section \ref{testing}, learned filters will change roles during testing.

The generalization of a single CNN element (neuron filter) to the entire network is mathematically cumbersome. We can make analysis on a subset of a network and then use generalized methods and theorems to understand what and how CNN learns. For that we analyze two neuron filters cascaded and show that how they learn is similar to a single unit. Two units are depicted in Figure \ref{ONE_TWO}. We use  S(.) for soft thresholding in following equations for simplicity without denoting biases. Take a patch $x_{k-2}$ with size $e\text{x}e$; $x_{k-1}$ with size $c\text{x}c$; $f_{k-2}$ and $f_{k-1}$, with size $a\text{x}a$.

\begin{align}
\label{cascadedsoft}
x_{k-1} &= S_1(x_{k-2}*f_{k-2})\notag
\\x_{k} &= S_1(x_{k-1}*f_{k-1})=S(S(x_{k-2}*f_{k-2})*f_{k-1})
\end{align}
Update of $f_{k-1}$ follows the same steps as discussed before. The update of $f_{k-2}$ requires derivation of mse w.r.t. $f_{k-2}$ where we will utilize the $X_{e,a}$ operator.

\begin{align}
mse &= \frac{1}{2}||\boldsymbol{t}-S_1(D_{L,k-1}\boldsymbol{f_{k-1}})||^2
\\&=\frac{1}{2}||t-S_1(X_{e,a}\boldsymbol{(f_{k-1}})S_2(D_{L,k-2}\boldsymbol{f_{k-2}}))||^2
\end{align}
To calculate gradient, define a modified dictionary 
\begin{equation}
\label{moddict}
P \doteq X_{e,a}(f_{k-1}) D_{L,k-2}
\end{equation}
Then we can calculate gradient of MSE. For readability issues we will stop using bold letters for vectors.
\begin{equation}
\frac{\partial mse}{\partial {f_{k-2}}} = -(P^T t - P^T S_1(X_{e,a}(f_{k-1})S_2(D_{L,k-2}f_{k-2}))) = \notag
\end{equation}
\begin{equation}
\label{twolayergradient}
\left\{
        \begin{array}{ll}
            - (P^T (t -  P f_{k-2}+ X_{e,a}(f_{k-1})b_{k-2}+ b_{k-1}))
           \\ \quad \quad \quad \quad if\  D_{L,k-2}f_{k-2} > b_{k-2} \ 
            \\ \quad \quad \quad \quad and \ if \ X_{e,a}(f_{k-1})(D_{L,k-2}f_{k-2}-b_{k-2}) > b_{k-1} 
           \\ -(P^T t) \quad o.w.
        \end{array}
    \right.
\end{equation}
Then we calculate
\begin{equation}
    {f_{k-2} ^{new}}={f_{k-2}}-\frac{\partial mse}{\partial {f_{k-2}}} = \notag
\end{equation}
\begin{equation}
\label{twolayergradient2}
\left\{
        \begin{array}{ll}
           {f_{k-2}}+P^T( {t} -  P {f_{k-2}}+ X_{e,a}({f_{k-1}}){b_{k-2}}+ {b_{k-1}})
           \\ \quad \quad \quad \quad if\  D_{L,k-2}{f_{k-2}} > {b_{k-2}} \ 
            \\ \quad \quad \quad \quad and \ if \ X_{e,a}({f_{k-1}})(D_{L,k-2}{f_{k-2}}-{b_{k-2}}) > {b_{k-1}} 
           \\ {f_{k-2}}+P^T {t} \quad o.w.
        \end{array}
    \right.
\end{equation}
We can also define a new equivalent function as in previous case, by choosing bias value as given in equation \ref{probably_not_used}.
\begin{equation}
\label{second}
f_{k-2} ^{new} \doteq S_{b'}(f_{k-2} +P^T (t -  P f_{k-2}))
\end{equation}
\begin{equation}
\label{probably_not_used}
b'=\left\{
        \begin{array}{ll}
          -X_{e,a}(f_{k-1})b_{k-2}-b_{k-1}
           \\ \quad \quad \quad \quad if\  D_{L,k-2}f_{k-2} > b_{k-2} \ 
            \\ \quad \quad \quad \quad and \ if \ X_{e,a}(f_{k-1})(D_{L,k-2}f_{k-2}-b_{k-2}) > b_{k-1} 
           \\=-Pf_{k-2} \quad o.w.
        \end{array}
    \right.
\end{equation}

Equation \ref{second} leads to the same equation as we have reached before in equation \ref{gradient}. From this point we can generalize that a CNN can solve for an inverse problem during training by an IST algorithm. The aim of a neuron filter, or the entire network, is to match its output to a target image to minimize error in equation \ref{oops2}. The estimation of neuron filters in this way is analogous to basis pursuit of K-SVD algorithm for sparse representation estimation. Therefore the vector f is a representation vector in equation \ref{oops2}. The dictionary, $D_L$, upon which representation of target, t, is found is constructed out of subpatches from low resolution image (superpatch). This seems only convenient because during testing (forward pass) the only information, from which the inverse problem is solved, is the input image itself. 

\begin{figure}
	\centering\includegraphics[width=1\linewidth]{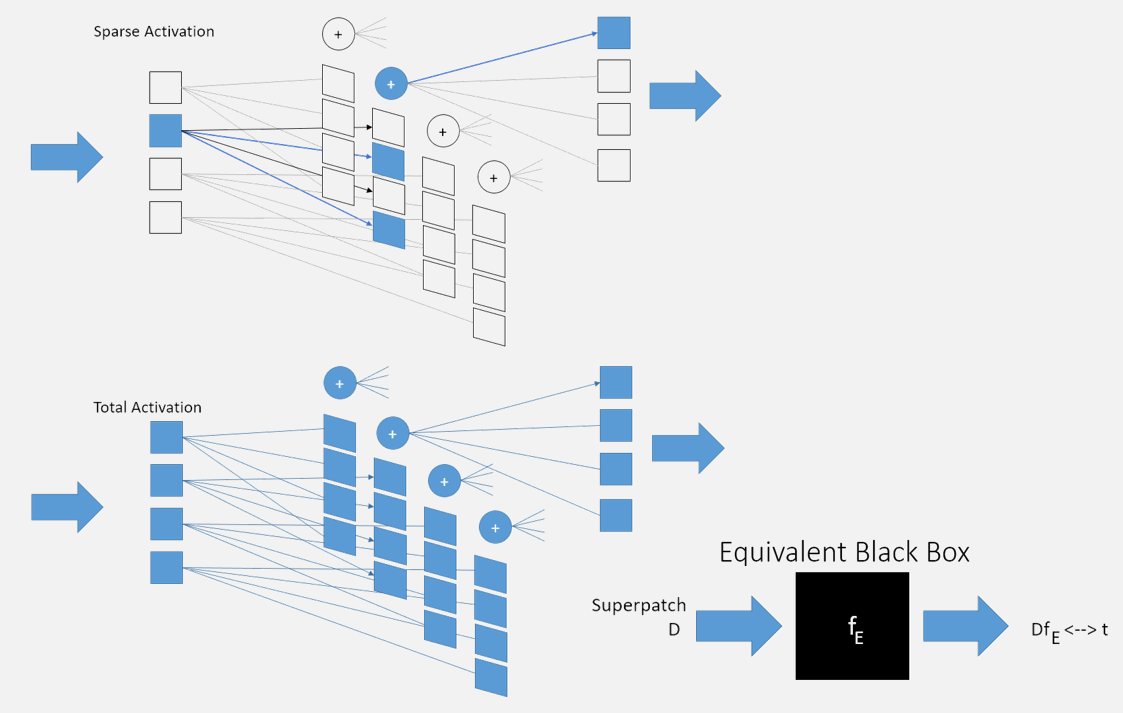}
	\caption{Illustating the Equivalent Filter $f_E$}
	\label{EQ}
\end{figure}

To further understand how a CNN learns and works we can inspect cases where mathematical proving is easier. Let us now recall the Landweber equation applied for CNN $f_E=(D_L^TD_L+\lambda I)^{-1}D_L^Tt$. This is illustrated in Figure \ref{EQ} In order to be able to use insights from this equation assume that all neurons in the network are activated for the inputs. For that unrealistic case, the network filters can be convolved among themselves to produce an end point filter, f$_E$. This is feasible because when all neurons are activated, their linear unit outputs are going to be the convolution results minus a bias that can be added up at the end, simply enabling the convolution of all filters to be applied in a single instant. A similar work is done by Mallat et. al. \cite{MALLAT} to analyze linearization, projection and separability properties of sparse representations for deep neural networks. 

The vector f$_E$ is going to be a normalized projection of t onto input image domain. Considering the rows of D$_L^T$ matrix, each row is a vectorized subpatch, thus each multiplication result from D$_L^T$t is going to be $\langle subpatch,t\rangle$ meaning the projection of target patch onto an input subpatch. D$_L^T$D$_L$ matrix have elements of inner products of subpatches such as $\langle subpatch_i,subpatch_j\rangle$. The diagonals of D$_L^T$D$_L$ matrix, therefore, are normed square of each subpatch. The inverted matrix is going to be mostly composed of diagonals that are inverted normed square values of subpatches. This means that the entire equation calculates the projection of target patch, t, onto the input image domain. In other words, the result, f$_E$, consists of scores which measure how similar t vector is to each subpatch from the entire superpatch. If the target image has content that cannot be recovered by using certain region of input image, the reconstructed image is going to be inferior. This is due to the fact that the inverse problem operates solely on an input image. Selection of a larger area for the reconstruction of certain target patches proves useful because of increased information included into the system.

This insight provides a method for determining how deep a network should be for certain features. For example when the superpatch and corresponding target region contains only irregular texture, which can be modeled as gaussian noise, the D$_L$ matrix becomes linearly independent, meaning easily invertible. Consequently when the training set consists solely of textured images, shallow networks will perform as good as deep networks. 

In general the training set contains various features with different variances. Therefore the generalization of the new concepts that are introduced here are difficult. Training with different structures enables the constant evolution of neuron filters during training. However to have an activating branch for each feature either the network should have increased number of filters or the network will not converge which can be explained by the manifold hypothesis, as representations not covering the high dimensional input space \cite{MANIFOLD}.

We propose separating a CNN into two separate networks that will be trained with different data. The data separation can be carried out in multiple ways, one of which is a method that is utilized by Romano et. al. \cite{RAISR} for a very similar purpose: learning different filters for SR and deblurring. The method uses gradients in x and y directions and calculates what we term as "spatial coherency" to distinguish from Papyan's dictionary mutual coherence. Gradient estimates for x and y directions are vectorized into two vectors. Two vectors are concatenated under a matrix G that has the size 2 by M. An Eigen analysis is carried out over the matrix G$^T$G. Since this is a 2x2 matrix eigenvalues and eigenvectors can be obtained easily. The normalized difference of larger eigenvalue $\sqrt{\lambda_1}$ and smaller eigenvalue $\sqrt{\lambda_2}$ can be seen as the spread of the local gradients which reveals information about spatial coherency. The spatial coherency value can be calculated as:

\begin{equation}
\label{spcoh} 
\mu _k = \frac{\sqrt{\lambda_1}-\sqrt{\lambda_2}}{\sqrt{\lambda_1}+\sqrt{\lambda_2}}
\end{equation}

Using $\mu_k$, structured patches and irregular patterns can be separated. Two distinct features in two training sets enables the specialization of CNN for the solution of an inverse problem. For high spatial coherency data, we can use a deeper network to increase receptive field, consequently to increase included information for the inversion of the observation model. For low spatial coherency images we can use more shallow networks since information content is already higher despite its complexity.

\subsection{Testing (Reconstruction) Phase}
\label{testing}
\begin{figure}
	\centering\includegraphics[width=1\linewidth]{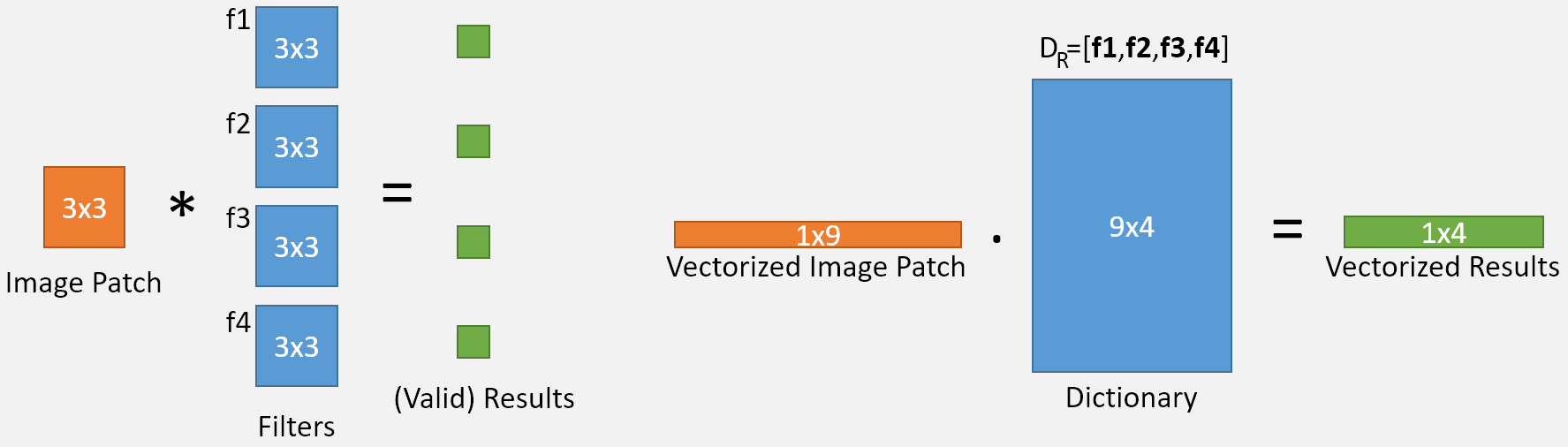}
	\caption{Formation of Reconstruction Dictionary ($D_R$)}
	\label{DictRec}
\end{figure}

For the testing phase, a new representation - dictionary duality (RDD) concept is proposed. RDD concept states that the representation vectors learned during the training phase can be used as atoms of a dictionary for the testing phase. The cost function that is minimized by CNN training (learning) yields a representation vector as the neuron filter. \textit{During testing (scoring, reconstruction) phase, resulting representation vectors (filters) from a layer of neurons turn into a dictionary (later named as $D_R$) upon which the reconstruction of output image is carried out.} We propose the idea that dictionaries and representations swap roles during training and testing. Also during training, inputs to each layer is perceived as a dictionary for the next layer. Following the idea of RDD, the neuron filter can be viewed as an atom of a dictionary consisting of many other neuron filters among the network layer. During testing period, the neuron filters are vectorized and concatenated to form the \textit{reconstruction} dictionary matrix D$_{R}=[\boldsymbol{f_1};\boldsymbol{f_2};\boldsymbol{f_3}...]$. A layer's output will be the \textit{representation} vector of input image in terms of the dictionary atoms, i.e. the neuron filters. This is illustrated in Figure \ref{DictRec}

To explain the representation problem analogy better we use ideas from two papers by Papyan et. al. \cite{PAPYAN16}\cite{PAPYAN17}. The authors have described a convolutional sparse coding problem where an observation is represented by a dictionary and its representation vector is further represented by another dictionary for a number of layers. They have proven, in Theorem 1, that layered representations can be estimated by a CNN based forward pass where mentioned layered dictionaries are filters of each layer of CNN. (Theorem 10 in original text \cite{PAPYAN16})

\textbf{Theorem 1}
	Suppose \textbf{g = y + n} where \textbf{n} is noise whose the power of noise is bounded by $\varepsilon_0$ and \textbf{y} is a noiseless signal. Considering a convolutional sparse coding (CSC) structure where $D_{R,l}$ is the dictionary, constructed from $l^{th}$ layer filters
	$\\ \ y=D_{R,1}x_1\\
	\ x_1=D_{R,2}x_2\\
	\ .\\
	\ .\\
	\ x_{N-1}=D_{R,N}x_N \\$
	Let {$\hat{x}_i$} be a set of solutions obtained by running a convolutional neural network, or layered soft thresholding algorithm with biases {$b_i$} as $\hat{x}_i=soft_i\{D_{R,i} ^T\hat{x}_{i-1}\} where \ \hat{x}_0=g$. Denote $|x_i ^{max}|$ and $|x_i ^{min}|$ as absolute maximum and minimum entries of $x_i$. Then assuming for $\forall 1\le i\le N\\$
	$||x_i||_0<\frac{1}{2}(1+\frac{1}{\mu(D_{R,i})} \frac{|x_i ^{min}|}{|x_i ^{max}|})-\frac{1}{\mu(D_{R,i})}\frac{\varepsilon_{i-1}}{|x_i ^{max}|}\\$
	where $\mu(D_{R,i})$ is the mutual coherence of the dictionary then
	
	1. The support of the solution $\hat{x}_i$ is equal to the support of $x_i$
	
	2. $||x_i - \hat{x}_i||_2 \le \varepsilon_i \\
	where\  \varepsilon_i=\sqrt{||x_i||_0}(\varepsilon_{i-1}+\mu(D_{R,i})(||x_i||_0 -1)|x_i ^{max}|+b_i)
	\\$

The mutual coherence $\mu(D_{R,i})$ is defined as $\underset{n\neq m}{min}|d_{R,i,n}^Td_{R,i,m}|$ where $d_{R,i,n}$ is $n^{th}$ column of $D_{R,i}$ \cite{PAPYAN16}. Although as stated by the Papyan et. al. there are tighter conditions for mutual coherence calculation. Later in this section we are going to analyze a different approach to the proving condition of this theorem.

Using Theorem 1, we can relate sparse representation problem with CNN forward pass. Elaborating further on separate networks in this context we can analyze CNNs better. Although image spatial coherency and dictionary (neuron filters') mutual coherence are two distinct measures they are correlated. We know that a high spatial coherency training set will yield learned filters of similar information content. This means that the neuron filters will be composed of mostly flat and singular orientation features thus filters already have higher mutual coherence. For low spatial coherency networks, the learned filters will exhibit textured features reducing the mutual coherence (reducing similarity) in the process. Theorem 1 is used to show the stability of CNNs for data representation. There are two different stability concepts that are proven by this theorem, one is having bounded response for small perturbations in the input data which is trivial in context of CNNs, the other more important concept of stability is the accuracy of the results given an input data. The theorem is valid, depending on mutual coherence of dictionary elements. The condition is given as
\begin{equation}
\label{cond}
||x_i||_0<\frac{1}{2}+\frac{1}{\mu(D_{R,i})} \frac{1}{2|x_i ^{max}|}(|x_i ^{min}|-2\varepsilon_{i-1})
\end{equation}

In this condition, at first glance it looks as $\mu(D_{R,i})$ is required to be as low as possible and $\frac{|x_i ^{min}|}{|x_i ^{max}|}$ as close to 1 as possible. But further analysis reveals that $\mu(D_{R,i})$ and $x_i ^{min}$, $x_i ^{max}$ values are interdependent, therefore a straightforward upper bound may not be defined. If mutual coherence decreases, $\underset{i\neq j}{min}|d_{R,i,n}^Td_{R,i,m}|$ decreases. It means the eigenvalue spread of $D_{R,i} ^T D_{R,i}$ increases. Given enough number of elements in a dictionary this means maximum and minimum values of $\hat{x}_i$ are also pushed apart. Assuming a stable solution exists, this means an increase in $|x_i ^{max}|$ and a decrease in $|x_i ^{min}|$, since difference of estimation and true value is bounded by the power of noise. The existence of a solution for noisy case CSC (Theorem 1) is not proven in Papyan et. al. \cite{PAPYAN16}. Therefore we have to look for conditions that would contradict with the existence of a non-trivial solution. For equation \ref{cond} to be valid for non-trivial solution, $\frac{1}{\mu(D_{R,i})} \frac{1}{2|x_i ^{max}|}(|x_i ^{min}|-2\varepsilon_{i-1})$ should be greater than $\frac{1}{2}$. Or if we define a looser but easier bound: 

\begin{equation}
\label{bound}
    |x_i ^{min}| > 2\varepsilon_{i-1}
\end{equation}

If we assume the existence of a stable solution we would have $||x_i - \hat{x}_i||_2 \le \varepsilon_i$ where $\hat{x}_i=soft_i\{D_{R,i} ^T\hat{x}_{i-1}\}$ then
\begin{align}
\label{zaa}
&|x_i ^{min}| = \underset{n}{min}|x_i[n]| \le \underset{n}{min}|\hat{x}_i[n]|+\sqrt{\varepsilon_{i}} 
\\ &\le \underset{n}{min}|d_{R,i,n} ^T\hat{x}_{i-1}|+\sqrt{\varepsilon_{i}}
\\ &\le \underset{n}{min}|d_{R,i,n} ^T x_{i-1}|+\sqrt{\varepsilon_{i}}+\sqrt{\varepsilon_{i-1}} 
\\ &= \underset{n}{min}|d_{R,i,n} ^T D_{R,i}x_{i}|+\sqrt{\varepsilon_{i}}+\sqrt{\varepsilon_{i-1}}
\\ &\le \underset{n \neq m}{min}|d_{R,i,n} ^T d_{R,i,m} |.||x_1||_2+\sqrt{\varepsilon_{i}}+\sqrt{\varepsilon_{i-1}}
\\ &=\mu(D_{R,i})||x_1||_2+\sqrt{\varepsilon_{i}}+\sqrt{\varepsilon_{i-1}}
\end{align}

To satisfy equation \ref{bound}
\begin{equation}
\label{mubound}
    \mu(D_{R,i}) > \frac{2\varepsilon_{i-1}-\sqrt{\varepsilon_{i-1}}-\sqrt{\varepsilon_{i}}}{||x_i||_2}
\end{equation}

If mutual coherence is not above a certain value (equation \ref{mubound}) then the theorem does not hold. When conditions are not satisfied, $||x_i - \hat{x}_i||_2<\varepsilon_i$ statement is no longer valid. 
Starting from $\hat{x}_0=g$, The estimate $\hat{x}_1$ will be perturbed beyond the power of noise of the input. Recall that we have defined the noiseless signal y as $y=D_{R,1}D_{R,2}..D_{R,N}x_N$ where N is the number of layers, or stages. As the estimate of each stage drifts further from the original representation values, in MSE sense, the end result of N layered CNN becomes an inaccurate representation of the original data.
\begin{equation}
\label{inaccuraterep}
    \hat{y}=D_{R,1}D_{R,2}..D_{R,N}\hat{x}_N
\end{equation}

Although estimating y, in forward problem that is defined as g=y+n, is a denoising problem, which is an example of an inverse problem nonetheless, the stability discussion can be applied to generalized inverse problem for imaging as $g=Kt+n$. Then the equation in the theorem changes as  $Kt=D_{R,1}D_{R,2}..D_{R,N}x_N$ and as CNN learns to invert the observation system K, $x_N = t$.

Skip connections relay information from previous layers to deeper layers. One must consider both training and testing to understand the effect of skip connections on mutual coherence of layer elements and the stability of CNN overall. During forward pass (testing), information from previous layers are added on deeper layers' outputs. As prior layers' information is more similar to the original data compared to deeper layers' information, Because of less perturbation caused by equation \ref{inaccuraterep}. Skip connections help preserving the accuracy of the end result. During backpropagation (training), the total error is carried, via skip connections, to deeper layers. Bypassing the chain rule of gradient calculation enables the true gradient to influence deeper filter coefficients. The initialization of CNN coefficients are done via Gaussian noise. Initial coefficients has the lowest mutual coherence possible. Backpropagating an error
that is calculated from difference of target data and input data that is lost within initial coefficients (as in modified dictionary in equation \ref{moddict}), without losing fidelity becomes almost impossible for deeper networks. This is the reason why the middle layers always have lowest mutual coherence (see Tables \ref{LCOHMC1} and \ref{LCOHMC2} in section \ref{exper}). Skip connections bypasses effect of many layers on the true gradient therefore skip connections help updating deeper layers with more accurate gradients. Consider the modified dictionary in equation \ref{moddict}. With skip connections the gradient descent will use direct input information $D_{L,k-2}$ instead of $ X_{e,a}(f_{k-1}) D_{L,k-2}$ that is dictionary modified with potentially unreliable $f_{k-1}$ at the initial stages of training. (To be more precise, with skip connection, gradient descent will use a mixture of $D_{L,k-2}$ and $P$). Thus with skip connections filter coefficients become more accurate in representing noiseless input data Kt, then output of CNN becomes more accurate in representing the target data t. \textbf{Therefore we have mathematically shown the benefit of skip connections between hidden layers or input-output layers as in residual learning \cite{VDSR} which provide the network with necessary mutual coherence.} Mao et. al. \cite{MAO} have proposed the usage of skip connections between convolution and deconvolution layers to prevent input information to be lost inside the encoder-decoder structure. Our reasoning for the usage of skip connections is to keep mutual coherence of filter coefficients high. This will provide the trained network to produce accurate results according to the error criterion used. During training, each layer's output is used as a dictionary for the next layer. The mathematical analysis of the learning process revealed in equation \ref{twolayergradient} that initial filters are updated with modified dictionaries $P$ that carry information of all the layers beyond it. Skip connections will reduce the variance of filters in different layers by carrying information across the network during forward and backward passes of training process. Although we have shown that CNNs benefit from skip connections exact structure is still subject to experimental refinement depending on the application.


\section{Experimental Validations}
\label{exper}
We have validated our assertions with experiments to solidify our results. We have used a superresolution example to test our findings. The training set was 291 images that Kim et. al. \cite{VDSR} have used. The test set was also from the same paper, which was Set5 and Set14 also BSD100. We separated the training set into two using spatial coherency value. The experimental findings are summarized as

\begin{itemize}
    \item Skip connections increase mutual coherence of middle layers, as well as other layers.
    \item Skip connections increase PSNR performance of the network
    \item Measured increase in values are subjected to T test and they are found statistically significant.
    \item Networks that are trained with lower spatial coherency data saturate in performance for shallower networks, while high spatial coherency data requires deeper network before converging in performance.
\end{itemize}

Tables \ref{LCOHMC1} and \ref{LCOHMC2} show that usage of Skip connection in a 10 layered network helped increasing mutual coherence. 

\begin{table}[]
\caption{Comparison of Mutual Coherence of Two Networks}
\label{LCOHMC1}
\begin{tabular}{|l|l|l|l|l|l|}
\hline
Layer \#    & 1      & 2       & 3      & 4      & 5        \\ \hline
Skip MC   & 0,0034 & 26e-5 & 0,0031 & 0,0034 & 99e-6 \\ \hline
NoSkip MC & 0,0016 & 18e-5 & 0,0026 & 0,0031 & 89e-6 \\ \hline
\end{tabular}
\end{table}

\begin{table}[]
\caption{Comparison of Mutual Coherence of Two Networks}
\label{LCOHMC2}
\begin{tabular}{|l|l|l|l|l|}
\hline
Layer \#    & 6       & 7      & 8      & 9      \\ \hline
Skip MC   & 18e-4  & 0,0026 & 0,0098 & 0,0038 \\ \hline
NoSkip MC & 99e-5 & 0,0013 & 0,0027 & 0,0030  \\ \hline
\end{tabular}
\end{table}

We have used 100,000 patches to test the PSNR of the results. The PSNR of the network without skip connections was averaged on 30.77 dB while bicubic result had 29.2 dB PSNR. The skip connections helped the PSNR to increase to 30.88 dB. We have used the so called T test to measure the significance of this result. T test is formulated as in equation

\begin{equation}
    T=\frac{mean_{set1}-mean_{set2}}{\sqrt{\frac{var_{set1}}{\#_{set1}}+\frac{var_{set2}}{\#_{set2}}}}
\end{equation}

where set1 contains the results of network with skip connection and set2 contains results from network without skip connection. T value is calculated to be 7.07 which points that the improvement is quite significant.

\begin{figure}
	\centering\includegraphics[width=1\linewidth]{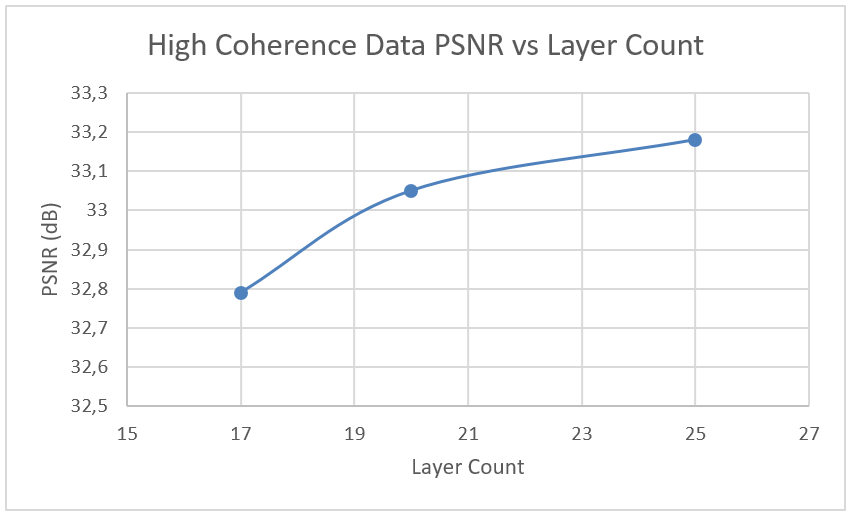}
	\caption{High Spatial Coherency Network Number of Layers Required for Performance Convergence}
	\label{HC}
\end{figure}

\begin{figure}
	\centering\includegraphics[width=1\linewidth]{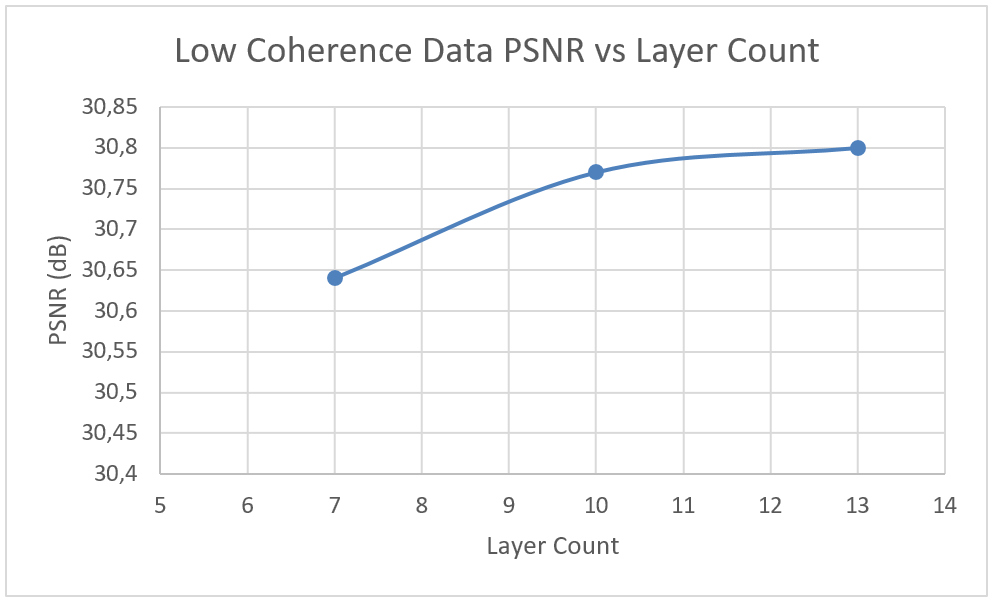}
	\caption{High Spatial Coherency Network Number of Layers Required for Performance Convergence}
	\label{LC}
\end{figure}

We have trained multiple networks for high spatial coherency and low spatial coherency data sets. As we have pointed out with our discussion in late section \ref{S:3}, High spatial coherency network required deeper networks before converging in performance, while low spatial coherency networks converged on shallower networks. This is illustrated in Figures \ref{HC} and \ref{LC}.

\section{Conclusion and Future Work}
Starting from the simplest element of a CNN we have brought mathematical clarification as to how CNN learning process works. We have proven that a neuron filter solves for an inverse problem during training. Then we have generalized the findings for convolutional networks. The result of training becomes a representation for the input image. We have discussed that after training process,these representation vectors, i.e. neuron filters, become feature selectors and dictionaries for reconstruction until the last layer. The CNN needs to satisfy conditions on mutual coherence between filters in each layer. These conditions ensure that the filters will be the true representations of their inputs during training. We have shown that to satisfy these conditions skip connections were necessary. We have discussed separation of a single network framework into two or more networks via spatial coherency information. This lead us to suggest different architectures for training with different content data. Low spatial coherency data inherently needs for more skip connections but also requires smaller receptive field thus less number of layers to converge to a solution.

Analyzing deep learning procedure is a complex problem. We have discussed simpler cases and proven key mathematical concepts on CNNs for inverse problems. In the future we will inspect effects of training with mini batches to the convergence of a network. Also we plan on analyzing optimizer with varying learning ratios and their effects on convergence.

\end{document}